\documentclass[twocolumn]{article}
\usepackage{graphicx} 
\usepackage{url} 
\usepackage{amsmath}
\usepackage{booktabs}
\usepackage{multirow}
\usepackage{setspace}
\usepackage[margin=1in, top=0.75in]{geometry}

\title{ \textbf{YOLOv11 Optimization for Efficient Resource Utilization}}

\author{Areeg Fahad Rasheed\textsuperscript{1},  Mahdi Zarkoosh\textsuperscript{2} \\
\textsuperscript{1}College of Information Engineering, Al-Nahrain University, Baghdad, Iraq \\
areeg.fahad@coie-nahrain.edu.iq \\
\textsuperscript{2}Independent Researcher, Baghdad, Iraq \\
mzarkoosh@gmail.com \\
}

\begin{document}

\date{}
\maketitle
\begin{abstract} \textbf{The objective of this research is to optimize the eleventh iteration of You Only Look Once (YOLOv11) by developing size-specific modified versions of the architecture. These modifications involve pruning specific layers and reconfiguring the main architecture of YOLOv11. Each proposed version is tailored to detect objects of specific size ranges, from small to large. To ensure proper model selection based on dataset characteristics, we introduced an object classifier program. This program identifies the most suitable modified version for a given dataset. The proposed models were evaluated on various datasets and compared with the original YOLOv11, YOLOv10, and YOLOv8 models. The experimental results highlight significant improvements in computational resource efficiency, with the proposed models maintaining the accuracy of the original YOLOv11. In some cases, the modified versions even outperformed the original model in terms of detection performance. Furthermore, the proposed models demonstrated reduced model sizes—each of the six proposed models showed a notable reduction compared to the original. Additionally, the required GFLOPs were reduced from 6.3MB (YOLOv11), 5.7MB (YOLOv10) and 8.1MB (YOLOv8) to just 3.8MB for the large model. All proposed models also achieved faster inference times, significantly reducing the time required to detect objects in images. Models weights and the object size classifier can be found in this repository \footnote{https://github.com/Zarko-ai/yolov11}}. 
\end{abstract}


\section{Introduction}

Object detection is one of the key tasks in computer vision. It represents the ability of a computer model to detect and classify visual objects in images. Object detection is widely used today in various applications in many fields, such as medicine \cite{ahmed2022detection, ragab2024comprehensive}, agriculture \cite{badgujar2024agricultural}, industry \cite{khanam2024comprehensive}, military \cite{haameid2021automatic}, etc.

Object detection has evolved through two significant milestones. In the early 1990s, it primarily relied on handcrafted features and sophisticated engineering due to limited image representation techniques \cite{viola2001rapid}. The second era began with the development of convolutional neural networks (CNNs) \cite{fukushima1980neocognitron,gu2018recent, rasheed2024unveiling}. However, the adoption of CNNs for object detection was initially delayed due to the lack of access to high-powered computational resources. As advancements in computational capabilities emerged, most object detection techniques in this era began to rely heavily on CNNs \cite{thompson2020computational}.

Object detection methods based on convolutional neural networks (CNNs) are typically classified into two types \cite{wu2020recent,zou2023object}: two-stage object detection and one-stage object detection. In two-stage object detection \cite{du2020overview}, the algorithm first scans the image to identify regions likely to contain objects using a Region Proposal Network (RPN) \cite{ren2016faster}. The RPN generates potential regions of interest (ROIs) by assigning an objectness score to each region and refining bounding box proposals. These proposals are then passed to a second stage, where a CNN predicts the object’s class and adjusts the bounding box coordinates. Popular algorithms in this category include R-CNN, Fast R-CNN, and Faster R-CNN \cite{ren2016faster,bharati2020deep}. This approach is known for its high accuracy and precision but is computationally intensive. In contrast, one-stage object detection skips the region proposal stage and directly predicts the classes and bounding boxes of objects in a single step \cite{zhang2021comprehensive}. This method integrates classification and localization in one unified network, making it faster and more suitable for real-time applications. A prominent example of this type is YOLO (You Only Look Once) \cite{redmon2016you}.

\begin{figure}
    \centering
    \includegraphics[width=1\linewidth]{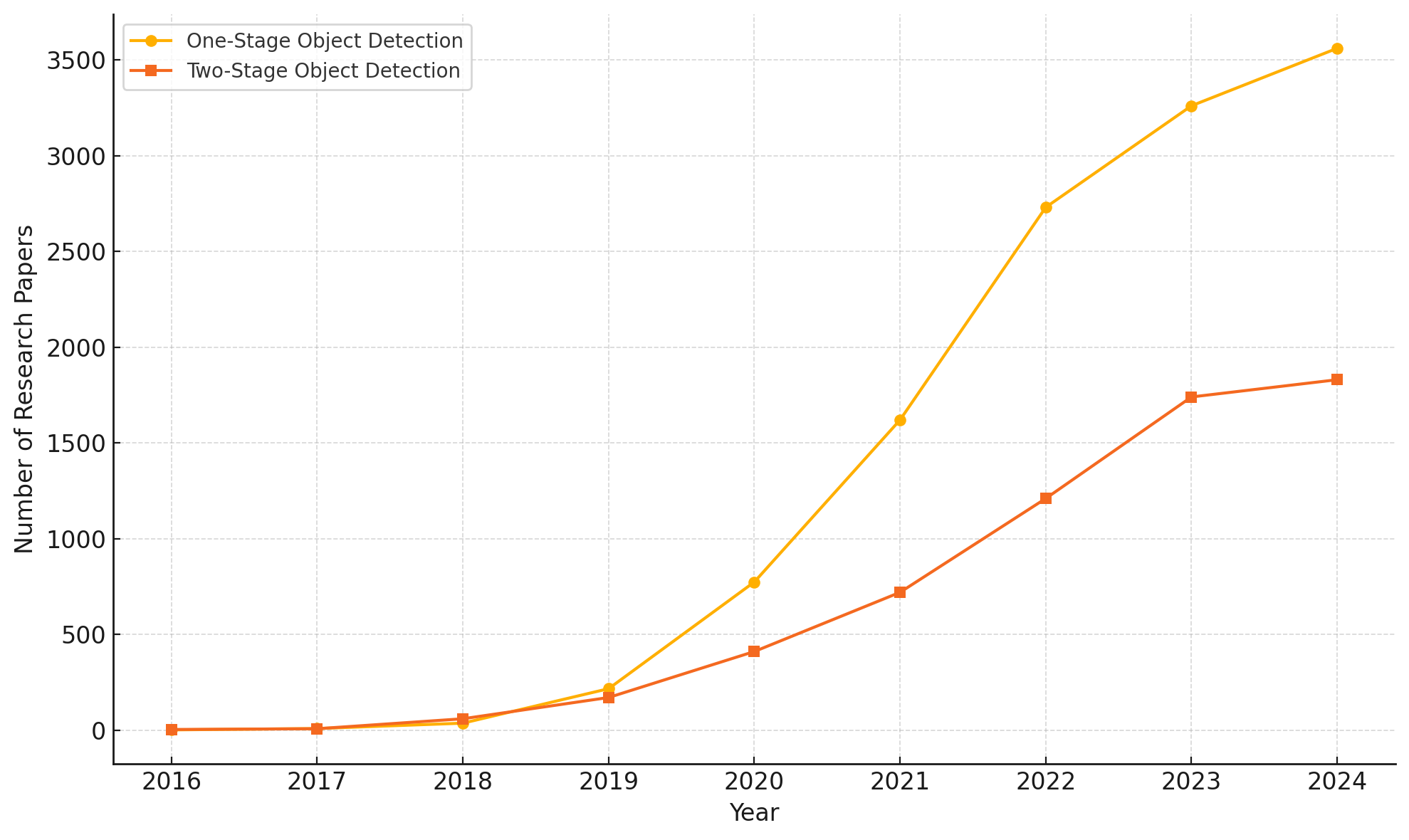}
    \caption{Showing the research trends for one-stage and two-stage object detection methods from 2016 to 2024. }
    \label{fig:paper_num}
\end{figure}

From 2016 to 2024, one-stage object detection methods have gained significantly more popularity and attention from researchers compared to two-stage methods. As illustrated in Figure \ref{fig:paper_num}, We present the number of research papers referencing one-stage and two-stage object detection during this period.

In this paper, we focus on a popular one-stage algorithm, YOLO—specifically version YOLOv11n, developed in 2024 \cite{yolo11, khanam2024yolov11}. The primary objective of this study is to propose six modified versions of YOLOv11n, each optimized for computational efficiency while preserving accuracy. Rather than utilizing the full YOLOv11n model, we adapt and tailor each version to detect objects of specific sizes more effectively. Although our proposed modifications are based on the n version, they can also be applied to the other YOLOv11 variants (s, m, x, l). The following are the main contributions of this work.


\textbf{Contributions}
\begin{enumerate}
    \item We proposed six modified versions of YOLOv11, targeting different object sizes. These models are based on modified YOLOv11 and are named as follows: YOLOv11-small, YOLOv11-medium, YOLOv11-large, YOLOv11-sm, YOLOv11-ml, and YOLOv11-sl.
    \item Each proposed model has been tested and evaluated on a relevant dataset. The dataset was selected based on an object classification program, and the results were evaluated using standard object detection metrics. The performance of the modified models was compared with the original YOLOv11, YOLOv10, and YOLOv8 Models.
    \item The proposed modified versions are applied to YOLOv11n and can also be adapted to other YOLOv11 variants, including YOLOv11s, YOLOv11m, YOLOv11l, and YOLOv11x.
\end{enumerate}

The rest of the paper is organized as follows: In Section 2, we provide an overview of the YOLOv11 model and its main components. In Section 3, we detail the proposed models. In Section 4, we describe the program used to classify the nature of the dataset and the dataset used to evaluate the models. In Section 5, we present the main results and comparisons with recent works. In Section 6, we discuss the limitations of the work and suggest future directions. Finally, in the last section, we conclude by summarizing the key findings.

\section{YOLOv11 Overview}
YOLO, created by Joseph Redmon et al, in \cite{redmon2016you}, is one of the prominent methods used for object detection. It is based on a one-stage approach, which processes the entire image in a single pass to predict bounding boxes and class probabilities \cite{hussain2024depth}. YOLO has undergone significant development, evolving from its first version to the latest (i.e., YOLOv11) version created by Ultralytics.

\begin{figure*}[htpb]
    \centering
    \includegraphics[width=1\linewidth]{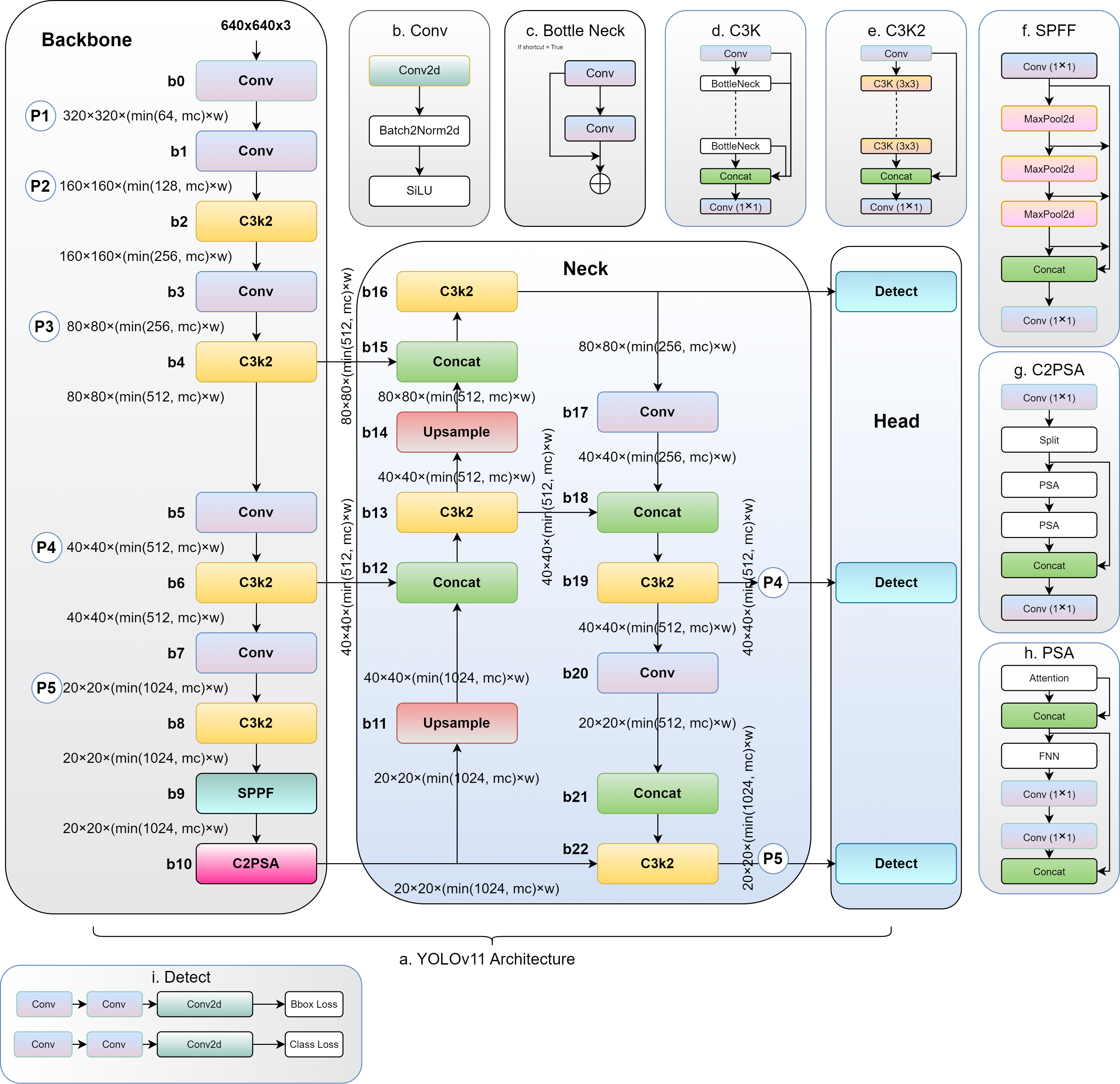}
    \caption{YOLOv11 architecture}
    \label{fig:yolov11}
\end{figure*}

YOLOv11 is the latest version of the YOLO family, offering significant improvements in speed, accuracy, and feature extraction. The architecture of YOLOv11, shown in Figure \ref{fig:yolov11}, highlights the main components of the model. It generally consists of three key components: the backbone, the neck, and the head. Below, we briefly illustrate each component and the features added to enhance the overall structure.

Backbone: The first main component of YOLOv11 is the backbone, which is responsible for extracting key features at different scales from the input image. This component consists of multiple convolutional (Conv) blocks, each containing three sub-blocks, as shown in Figure \ref{fig:yolov11}(b): Conv2D, BatchNorm2D, and the SiLU activation function. In addition to the Conv blocks, YOLOv11 includes multiple C3K2 blocks, which replace the C2f blocks used in YOLOv8 \cite{learnopencv}. The C3K2 blocks provide a more computationally efficient implementation of Cross-Stage Partial (CSP) \cite{wang2021scaled}, as illustrated in Figure \ref{fig:yolov11}(e). The final two blocks of the backbone are the Spatial Pyramid Pooling Fast (SPPF) and the Cross-Stage Partial with Spatial Attention (C2PSA) \cite{he2015spatial,khanam2024yolov11}. The SPPF block utilizes multiple max-pooling layers, as shown in Figure \ref{fig:yolov11}(f), to extract multi-scale features from the input image efficiently. On the other hand, the C2PSA block incorporates an attention mechanism as shown in figure \ref{fig:yolov11} (g), to enhance the model's accuracy.

Neck: The second main component of YOLOv11 is the neck. As shown in Figure \ref{fig:yolov11}, the neck consists of multiple Conv layers, C3K2 blocks, Concat operations, and upsample blocks, along with the advantages of the C2PSA mechanism. The primary role of the neck is to aggregate features at different scales and pass them to the head blocks \cite{learnopencv}. 

Head: The final component of YOLOv11 is the head, a crucial module responsible for generating predictions. It determines the object class, calculates the objectness score, and accurately predicts the bounding boxes for identified objects \cite{jegham2024evaluating}.

\section{Modified Versions of YOLOv11}

The backbone of the YOLOv11 architecture performs multiple down-sampling operations on the input image, reducing it to scales of 2x, 4x, 8x, 16x, and 32x. This process generates five sets of features (320x320, 160x160, 80x80, 40x40, and 20x20). These feature sets which are named as (P1, P2, P3, P4, P5) see figure \ref{fig:yolov11}, are combined with other components of the model, such as SPPF and C2PSA, and then fed into the head blocks. The larger feature sets are responsible for detecting large objects, while the medium-sized feature sets, such as 40x40, are used to detect medium objects. The smallest feature sets, such as 20x20, focus on detecting small objects \cite{alif2024yolov11,feng2024improved}.  

According to the official documentation, the YOLOv11 model's head contains three detection blocks, each responsible for detecting objects of specific sizes. For example, small objects are typically those with a size less than $32^2$ pixels, medium objects are those with a size greater than $32^2$ but less than $96^2$ pixels, and large objects are those with a size exceeding $96^2$ pixels \cite{ultralytics2024}.

In some cases, object detection applications are designed to focus on specific object sizes. For instance, aerial applications typically involve detecting small objects in images \cite{wang2021tiny,liu2020uav}. To enhance resource efficiency, instead of using the standard YOLOv11 architecture, we propose six modified versions of YOLOv11 tailored to detect specific object sizes. (YOLOv11-small, YOLOv11-medium, YOLOv11-large,  YOLOv11-sm, YOLOv11-ml, and YOLOv11-sl). Each model targets a specific object size, and the appropriate model is selected based on the datasets' objects size see Table \ref{tab:object-size}.

To streamline this process, we used a simple program to analyze and provide detailed information about the object sizes in the dataset \footnote{https://github.com/Zarko-ai/yolov11}, as discussed in the dataset section. Using these modified versions of YOLOv11 instead of the original architecture reduces the computational cost and model size while maintaining accuracy in most scenarios. 

In the following sections, we present each proposed modification of YOLOv11 and highlight the main differences between these models and the original architecture.

\begin{table*}[ht]
    \centering
    \caption{Object Size Categories for Modified YOLOv11 Models. Each model is optimized to detect specific object sizes based on the relative area to the image.}
    \begin{tabular}{|c|c|}
    \hline
    \textbf{Model Name}  & \textbf{Object Size Range} \\ \hline
    YOLOv11-small        & $\text{area} \leq 32^2$ \\ \hline
    YOLOv11-medium       & $32^2 < \text{area} \leq 96^2$ \\ \hline
    YOLOv11-large        & $\text{area} > 96^2$ \\ \hline
    YOLOv11-sm           & $\text{area} \leq 96^2$ \\ \hline
    YOLOv11-ml           & $32^2 < \text{area}$ \\ \hline
    YOLOv11-sl           & $\text{area} \leq 32^2 \text{ or } \text{area} > 96^2$ \\ \hline
    \end{tabular}
    \label{tab:object-size}
\end{table*}

\subsection{YOLOv11-small}

\begin{figure}
    \centering
    \includegraphics[width=0.7\linewidth]{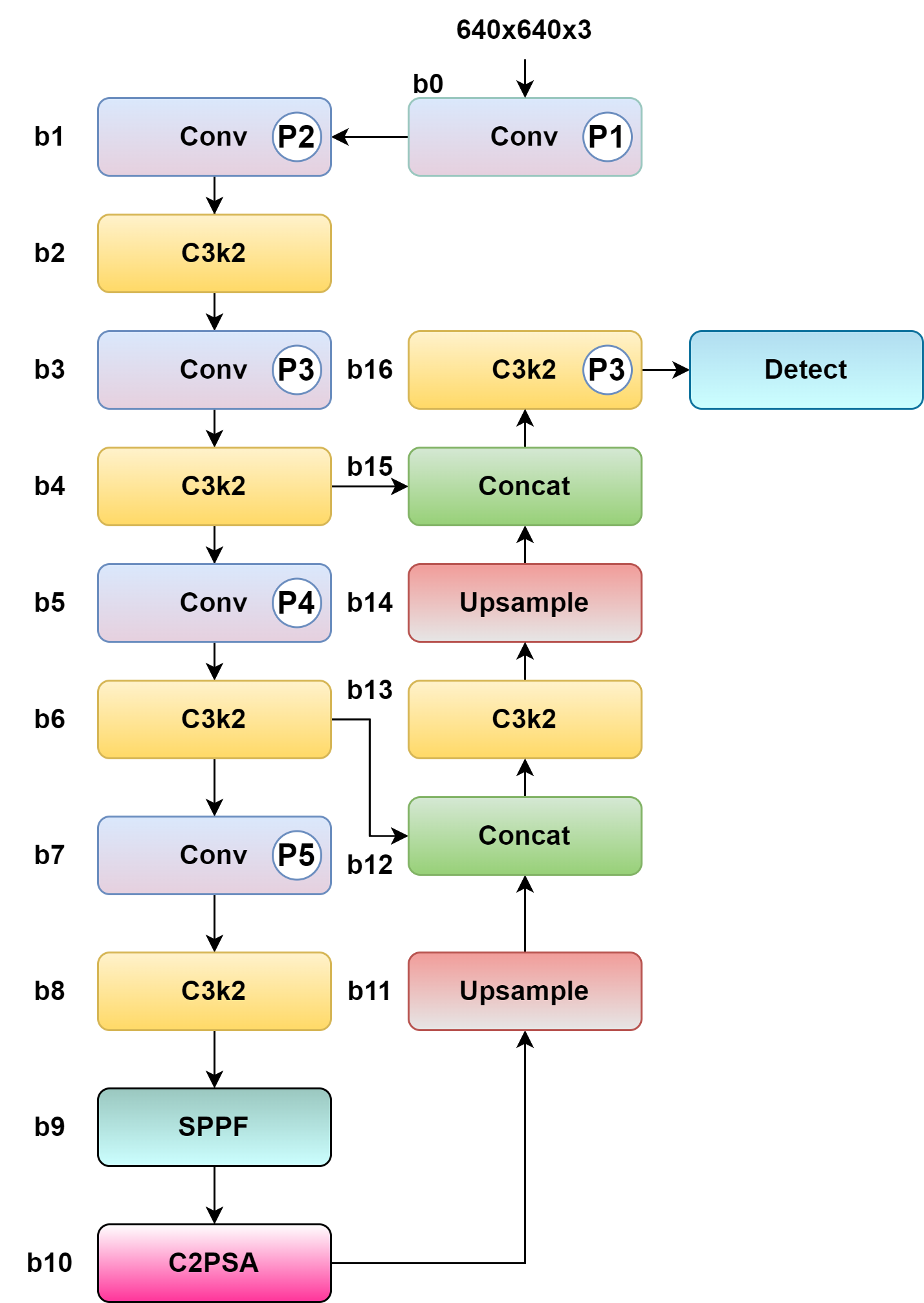}
    \caption{YOLOv11-Small Architecture for Small Object Detection}
    \label{fig:yolov11-small}
\end{figure}

The first modified version of YOLOv11 is the small version, which is designed to detect objects with an area less than or equal to \( 32^2 \). To modify YOLOv11, we have labeled each block of the original architecture starting with "b," ranging from b0 to b22 for simplicity. As mentioned in the previous section, the first detection head is used for detecting small object sizes \cite{alif2024yolov11, feng2024improved}. For this, we removed the second and third detection blocks. Since we removed these two detection blocks, we also eliminated the blocks that fed them with features important for detecting larger sizes. As a result, the blocks from b17 to b22, which are related to medium and large objects, were removed. The new version of YOLOv11, called YOLOv11-small, is shown in Figure \ref{fig:yolov11-small} .

\subsection{YOLOv11-medium}
\begin{figure}
    \centering
    \includegraphics[width=0.85\linewidth]{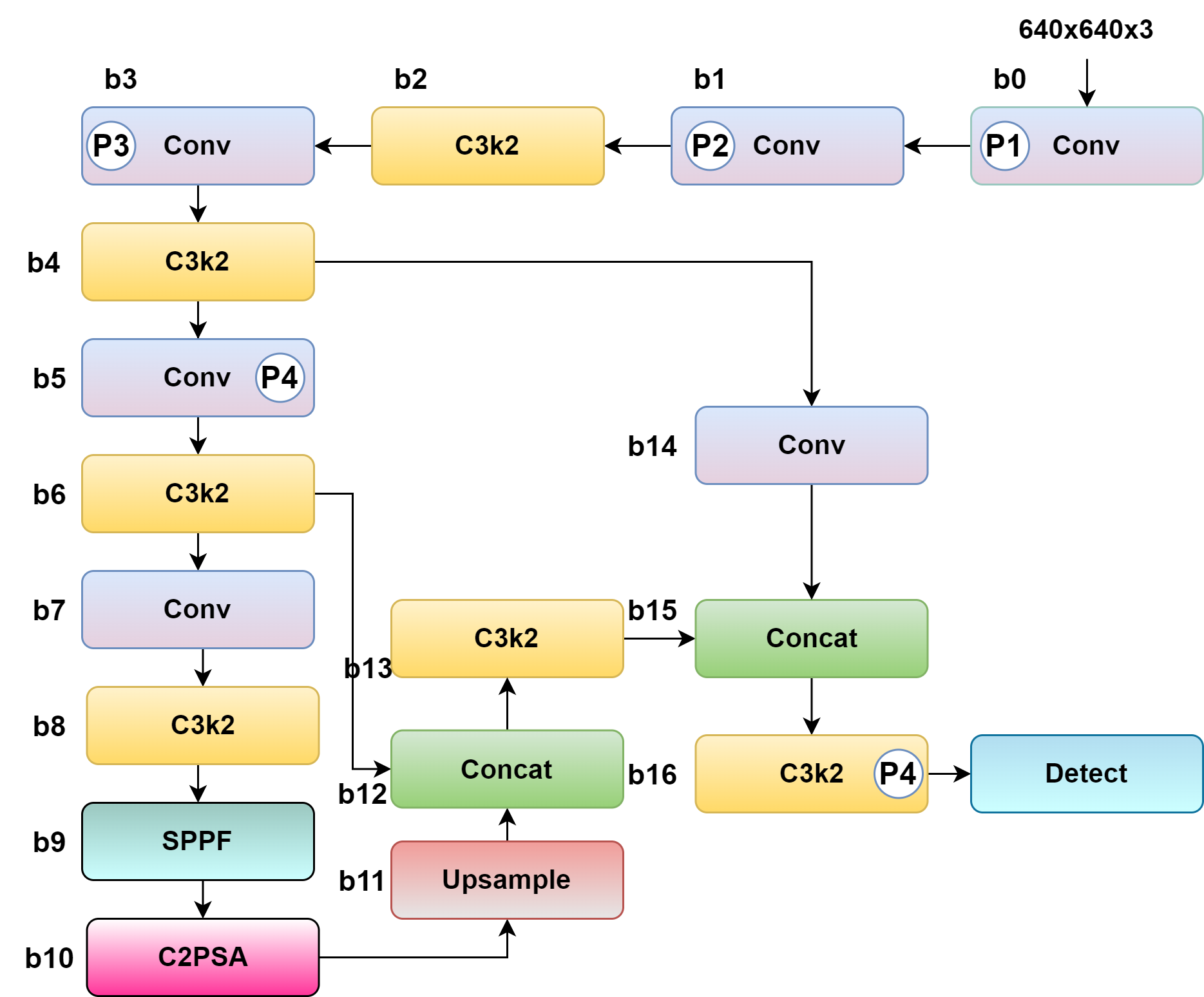}
    \caption{YOLOv11-Medium Architecture for Medium Object Detection}
    \label{fig:yolov11-medium}
\end{figure}

The second modified version of YOLOv11 that we proposed specifically targets medium-sized objects, defined as those with sizes greater than \(32^2\) and less than \(96^2\). Back to Figure \ref{fig:yolov11}, we removed all blocks related to detecting small and large objects. We eliminated the blocks responsible for processing features related to small and large object detection. Specifically, blocks \(b14\), \(b15\), and \(b16\) were removed as they feed detection heads for small objects \cite{alif2024yolov11, feng2024improved}. Similarly, blocks \(b20\), \(b21\), and \(b22\) were removed as they feed detection heads for large objects. After removing these blocks, we renamed the original YOLOv11 blocks associated with medium-sized objects (previously \(b17\), \(b18\), and \(b19\)) to \(b14\), \(b15\), and \(b16\), respectively, as shown in Figure \ref{fig:yolov11-medium}.

\subsection{YOLOv11-large}
\begin{figure}
    \centering
    \includegraphics[width=0.85\linewidth]{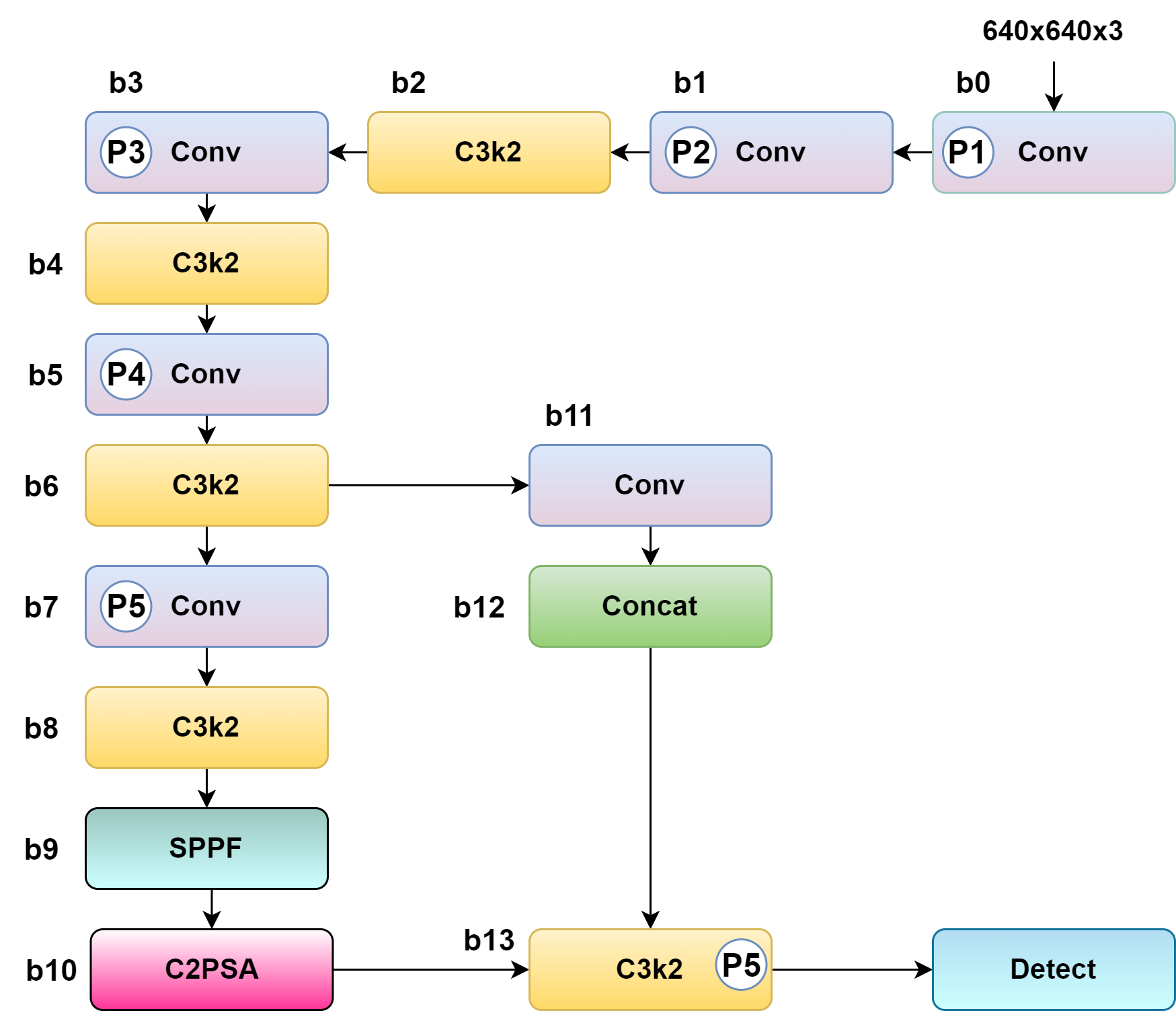}
    \caption{YOLOv11-Large Architecture for Large Object Detection}
    \label{fig:yolov11-large}
\end{figure}

The third modified version of YOLOv11 is designed to target large objects, where the object size area is greater than \(96^2\). To create the YOLOv11-large model, modifications were made to the original architecture by removing components unrelated to large object detection and reconnecting previously unlinked blocks \cite{alif2024yolov11, feng2024improved}. Specifically, blocks \(b11\) to \(b19\) were removed, as they are associated with providing features for detecting small and medium-sized objects. To ensure continuity in the network, block \(b19\) was reconnected to receive features from block \(b6\), as both utilize the same feature map as input. Additionally, blocks \(b19\), \(b21\), and \(b22\) were renamed to \(b11\), \(b12\), and \(b13\), respectively. The updated architecture of YOLOv11 for large objects is shown in Figure \ref{fig:yolov11-large}.

\subsection{YOLOv11-sm}
\begin{figure}
    \centering
    \includegraphics[width=0.7\linewidth]{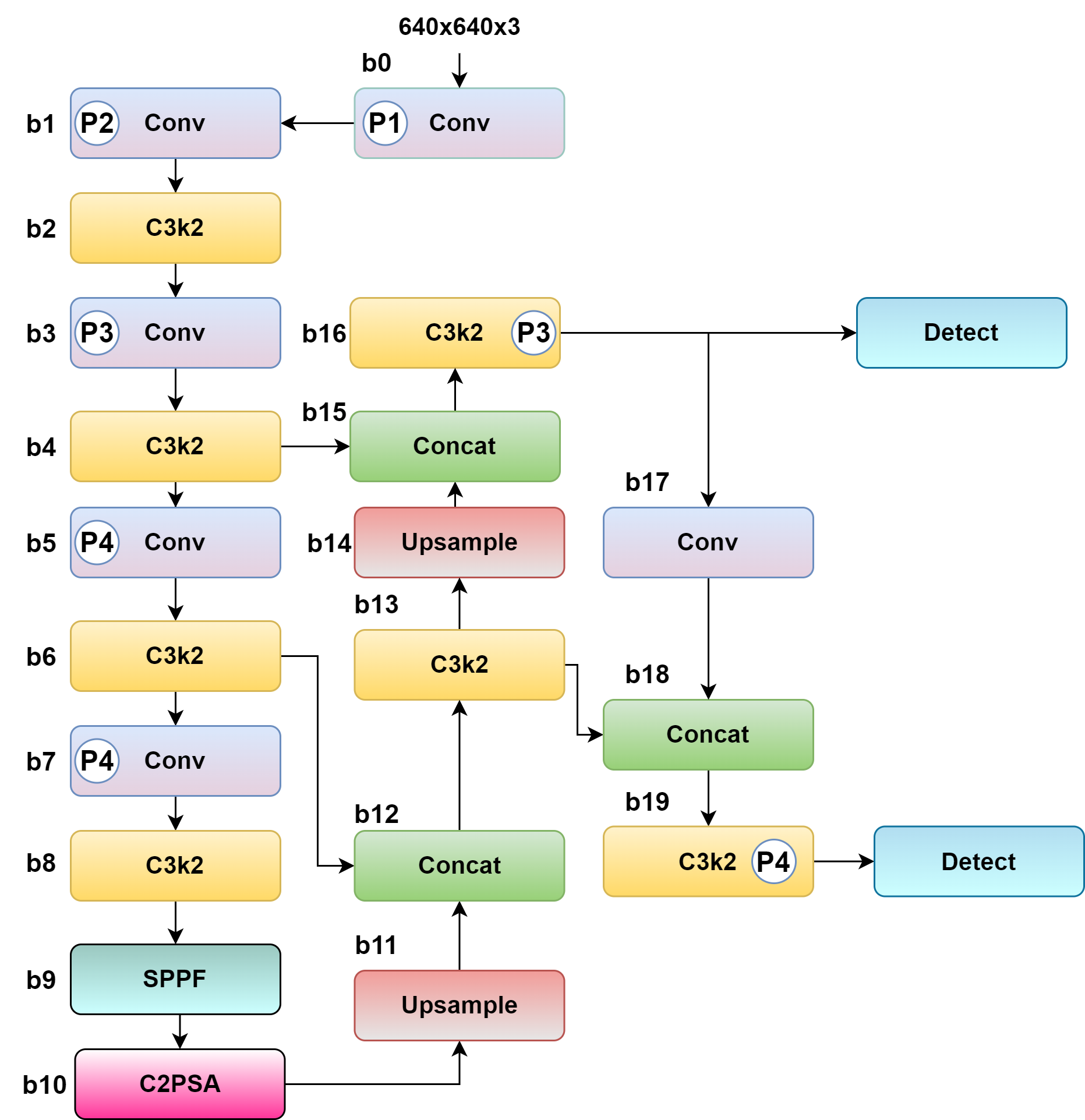}
    \caption{YOLOv11-sm Architecture for Small and Medium Object Detection}
    \label{fig:yolov11-sm}
\end{figure}

The fourth modified version of YOLOv11, named YOLOv11-sm, is specifically designed to provide flexibility for detecting small and medium-sized objects, where object size areas are less than \(96^2\), as shown in Table \ref{tab:object-size}. To implement this modified version, the blocks related to large object detection were removed, while the blocks for small and medium object detection were remained \cite{alif2024yolov11, feng2024improved}. As illustrated in Figure \ref{fig:yolov11}, the modification involves removing the third detection head and all the blocks that feed into it. Blocks \(b20\), \(b21\), and \(b22\) were removed, while the blocks from \(b0\) to \(b19\) remained unchanged. The updated architecture for small and medium object detection in YOLOv11-sm is depicted in Figure \ref{fig:yolov11-sm}.

\subsection{YOLOv11-ml}
\begin{figure}
    \centering
    \includegraphics[width=0.75\linewidth]{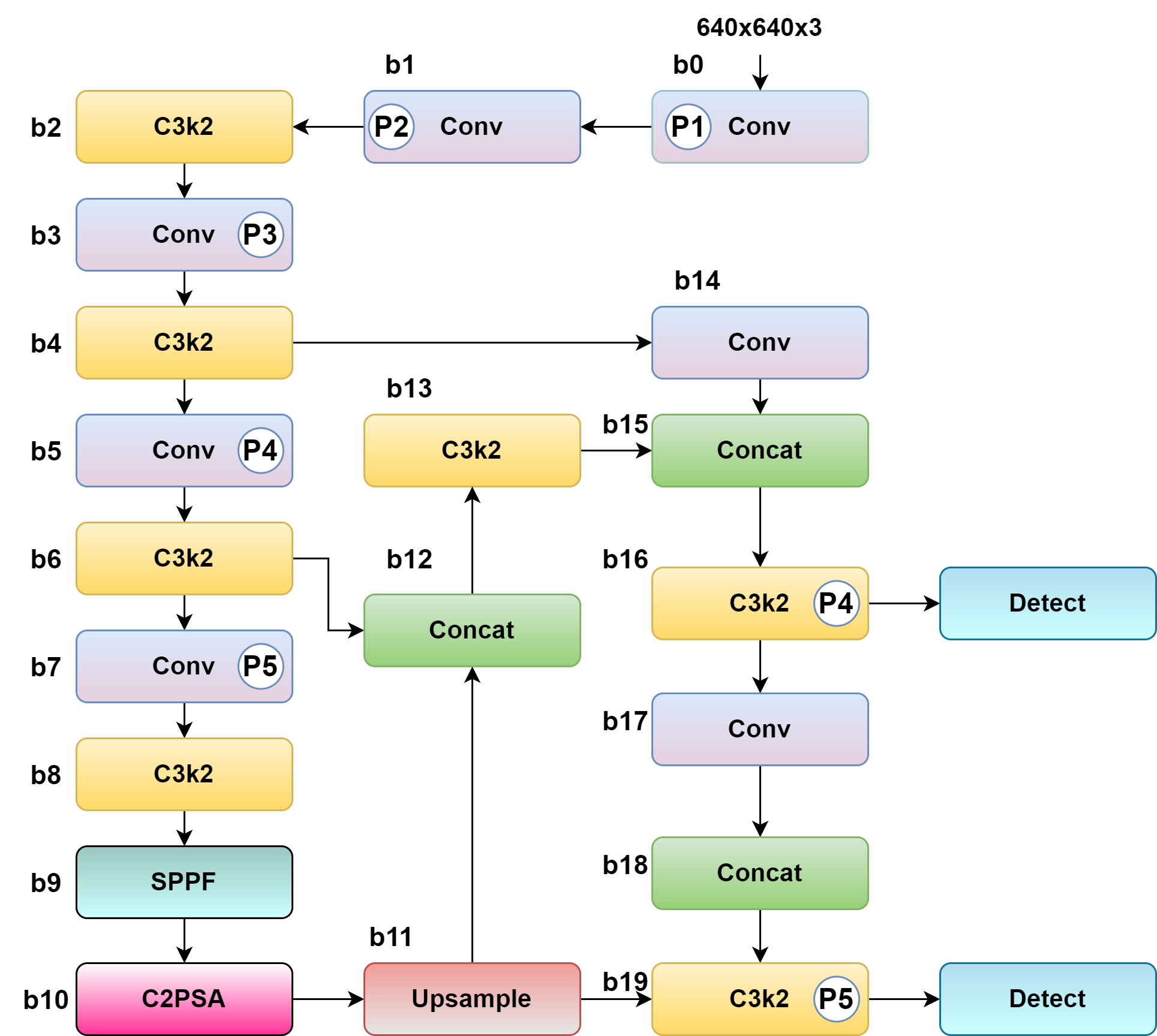}
    \caption{YOLOv11-ml Architecture for Medium and Large Object Detection}
    \label{fig:yolov11-medium-large}
\end{figure}

The fifth modified version of YOLOv11, referred to as YOLOv11-ml, is designed to target medium and large objects, with object sizes greater than \(32^2\). This means that any object exceeding this area can be effectively processed using this model. To derive YOLOv11-ml, modifications were made to the blocks related exclusively to small object detection. Specifically, as shown in Figure \ref{fig:yolov11}, blocks \(b14\), \(b15\), and \(b16\) were removed. Subsequently, the remaining blocks, \(b17\) to \(b22\), were renumbered as \(b14\) to \(b19\). Finally blcok number b4 linked with block b14. The updated architecture for medium and large object detection is presented in Figure \ref{fig:yolov11-medium-large}.

\subsection{YOLOv11-sl}
\begin{figure}
    \centering
    \includegraphics[width=0.75\linewidth]{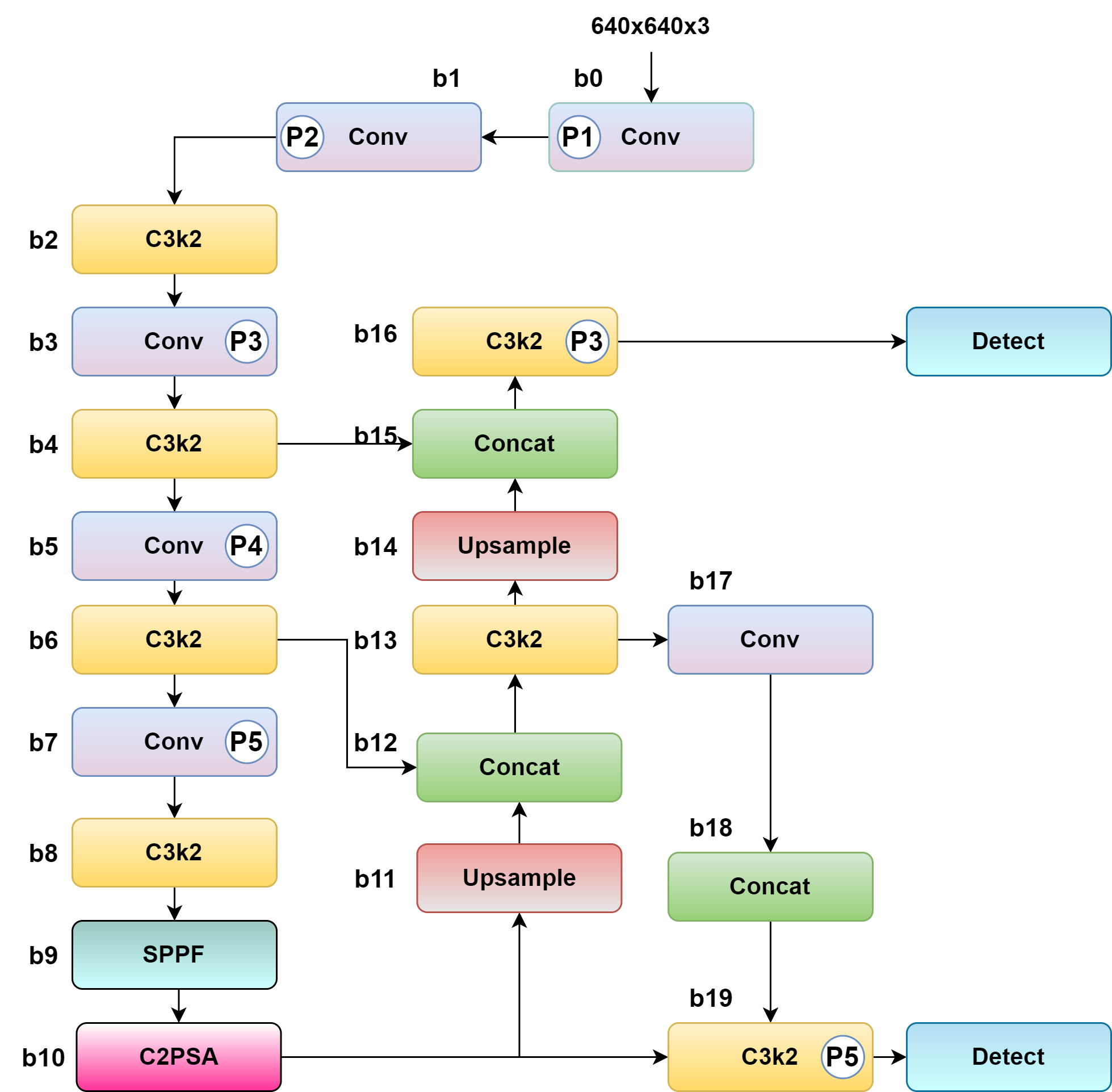}
    \caption{YOLOv11-sl Architecture for Small and Large Object Detection}
    \label{fig:yolov11-small-large}
\end{figure}

The sixth proposed modified version of YOLOv11, referred to as YOLOv11-sl, targets objects that are either small or large, as defined in Table \ref{tab:object-size}, where the object size must satisfy \(\text{area} \leq 32^2 \text{ or } \text{area} > 96^2\). To implement this model based on YOLOv11, the blocks related to medium-sized object detection were removed. Specifically, blocks \(b17\), \(b18\), and \(b19\) were eliminated, and block \(b20\) was reconnected to replace block \(b17\). The numbering of blocks was subsequently adjusted: blocks \(b0\) to \(b16\) remained unchanged, while blocks \(b20\), \(b21\), and \(b22\) were renumbered to \(b17\), \(b18\), and \(b19\), respectively. The architecture of this modified version is shown in Figure \ref{fig:yolov11-small-large}.

\section{Datasets Used and Object Size Classification Program}

We utilized six different datasets to evaluate the performance of the proposed models. These datasets were sourced from various environments, such as agriculture, medical fields, and others. In addition to the datasets, we also provide details of the program used for classifying the object sizes within the datasets to select the appropriate model for each scenario.

\subsection{Object size classifier}
As mentioned in the previous section, we proposed six modified versions of YOLOv11, each targeting different object sizes, as shown in Table \ref{tab:object-size}. To select the appropriate model for training, it is essential to understand the nature of the dataset being used. For this purpose, we developed a simple program that processes the training set of the dataset and estimates the area of all object sizes. Based on these measurements, the program helps select the most suitable model.

The program is used to analyze the nature of the dataset by computing the size of each object based on Table \ref{tab:object-size}. The size of each object instance is calculated from the label files and then classified according to the criteria defined in Table \ref{tab:object-size}.

The program workflow is as follows: it prompts the user to select the training label directory, iterates through all label files, and computes the size of each object. Each object size is then compared with the predefined size ranges specified in Table \ref{tab:object-size}. If an object instance is classified as small, the program increments the count for small objects, and similarly for other size categories. This process is repeated for all object instances across the dataset. Table 2 presents the object sizes for each dataset discussed in Section 4.2, along with the number of object instances corresponding to each size category.

\begin{table*}
    \centering
    \caption{Dataset Characteristics and the Proper Models for Evaluation}
    \label{table:dataset-properties}
    \begin{tabular}{|l|c|c|c|c|l|}
        \hline
        Dataset Name & Total Objects & Small & Medium & Large & Model Used \\ \hline
        WeedCrop \cite{sudars2020dataset} & 18,693 & 15,237 & 3,363 & 93 & YOLOv11-small \\ \hline
        BCCD \cite{bccd_dataset, kutlu2020white}& 11,780 & 547 & 10,094 & 1,139 & YOLOv11-medium \\ \hline
        Underwater Pipes \cite{ciaglia2022roboflow, underwater-pipes-4ng4t_dataset} & 12,238 & 4 & 551 & 11,683 & YOLOv11-large \\ \hline
        Aerial Airport \cite{aerial-airport_dataset} & 11,731 & 9,008 & 2,682 & 41 & YOLOv11-sm \\ \hline
        Brain Tumor \cite{braintumorataset,lin2024yolov8} & 21,526 & 1,056 & 3,484 & 16,985 & YOLOv11-ml \\ \hline
        Face Detection \cite{facedataset} & 620 & 21 & 0 & 599 & YOLOv11-sl \\ \hline
    \end{tabular}
\end{table*}

\subsection{Dataset used}

\begin{figure}
    \centering
    \includegraphics[width=0.995\linewidth]{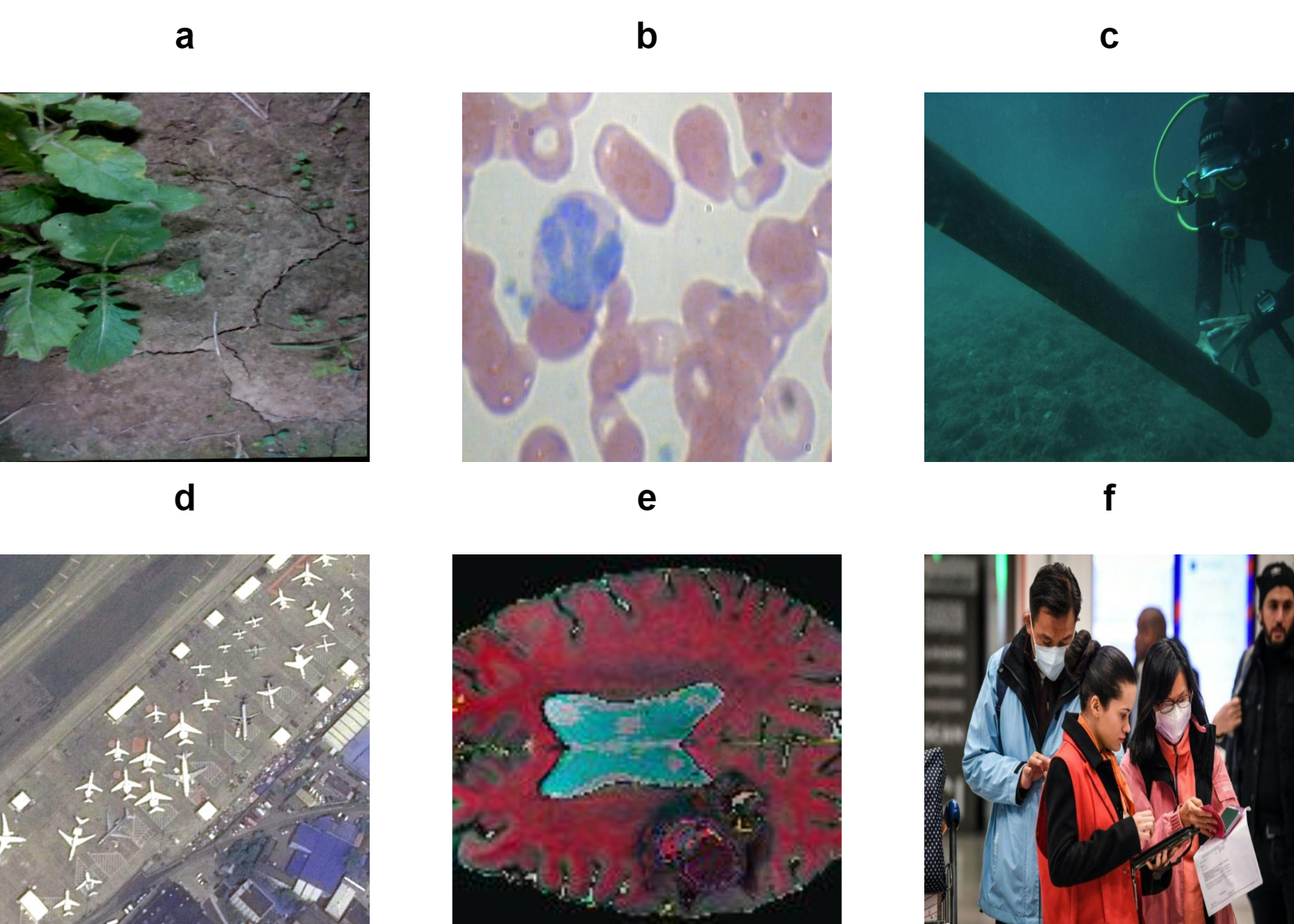}
    \caption{Representative samples from the datasets used — (a) WeedCrop, (b) BCCD, (c) Underwater Pipes, (d) Aerial Airplanes, (e) Brain Tumors, and (f) Face Detection.}
    \label{fig:dataset-sample}
\end{figure}

The following is a brief description of each dataset used in this paper for the model evaluation process. Figure \ref{fig:dataset-sample} shows sample form each used dataset. 

\begin{itemize}
    \item WeedCrop \cite{sudars2020dataset}: This dataset is used to detect and classify weeds and crops in the agricultural field. All objects in this dataset are categorized into two classes: weed or crop. It comprises 1,176 images containing approximately 7,853 objects. Augmentations such as rotations, shearing, and brightness adjustments were applied to expand the dataset, increasing the total number of objects to 18,693. According to Table \ref{table:dataset-properties}, most object instances in this dataset are categorized as small-sized. Therefore, this dataset will be used in the following section to evaluate the YOLOv11-small model.
    
    \item BCCD \cite{bccd_dataset, kutlu2020white}: The BCCD dataset is the second dataset used for classifying blood cells. It comprises 364 images with approximately 4,888 objects, the image size were 416x416, This dataset is commonly used in the medical field. The objects belong to three main classes: 'Platelets,' 'RBC,' and 'WBC.' Augmentation techniques, including flips, rotations, cropping, and adjustments to hue, saturation, brightness, and exposure, expanded the dataset to 874 images with a total of 11,780 objects. As shown in Table \ref{table:dataset-properties}, approximately 86\% of all object instances in the BCCD dataset are classified as medium-sized according to Table \ref{tab:object-size}. Therefore, this dataset will be used to evaluate the medium variant of YOLOv11 (i.e., YOLOv11-medium).
    
    \item Underwater pipes \cite{ciaglia2022roboflow, underwater-pipes-4ng4t_dataset}: The underwater pipes dataset is used for detecting pipes in underwater environments. It consists of a single class and contains a total of 7,971 images, with 12,238 object instances across the entire dataset.The large modified version of YOLOv11 will be used for this dataset, as the majority of object instances, approximately 96\% of the total objects, are classified as large, see Table \ref{table:dataset-properties}.
    
    \item Aerial Airport \cite{ciaglia2022roboflow, aerial-airport_dataset}: The Aerial Airport dataset is used for detecting airplanes in aerial environments. It consists of 338 images with 5,084 object instances. Augmentation techniques, including flips, cropping, and adjustments to hue, saturation, brightness, and exposure, expanded the dataset to 810 images with 11,731 object instances. The appropriate model for this dataset is YOLOv11-sm, as indicated in Table \ref{table:dataset-properties}, where most object instances in the images belong to the small and medium-sized categories.

    \item  Brain Tumor \cite{braintumorataset,lin2024yolov8}: The primary purpose of this dataset is to detect brain tumors. It comprises 9,900 images categorized into three classes. Across all splits—training, testing, and validation—the dataset contains a total of 21,526 object instances. Approximately 95\% of the object instances in the Brain Tumor dataset fall within the medium and large size ranges, as shown in Tables \ref{tab:object-size} and \ref{table:dataset-properties}. Therefore, the YOLOv11-sm modified version is used for the evaluation process of this dataset.
    
    \item Face Detection \cite{facedataset}: The final dataset used in this paper is the Face Detection dataset, which consists of 389 images, each containing multiple face instances, with a total of 620 faces.  The model used for this dataset is YOLOv11-sl, designed for both small and large objects, as most object instances in the dataset fall within these two size ranges.
\end{itemize}

\section{Evaluation of Modified YOLOv11 Models}

In this section, we present the performance of the proposed models and compare their results with the original YOLOv11 and YOLOv8 models. The main configurations parameters are present in Table \ref{tab:model-config}. The evaluation is divided into two main groups:
\begin{itemize}
    \item[1-] Accuracy and Detection Metrics: This group evaluates performance in terms of recall, precision, and mAP@50.
    \item[2-] Model Efficiency and Resource Utilization: This group focuses on model size, power consumption, inference time.
\end{itemize}

\begin{table}[h!]
\centering
\caption{Model Configuration Parameters}
\label{tab:model-config}
\begin{tabular}{|l|l|}
\hline
\textbf{Parameter}      & \textbf{Value}       \\ \hline
Epochs                 & 150                  \\ \hline
Seed                   & 0                    \\ \hline
Batch Size             & 16                   \\ \hline
Weight Decay           & 0.0005               \\ \hline
Patience               & 100                  \\ \hline
Learning Rate          & 0.01                 \\ \hline
\end{tabular}
\end{table}

\subsection{Accuracy and Detection Metrics}
In this section, we evaluate the model performance using metrics such as Recall, Precision, mAP@50. Four models have been used for the evaluation: the first is the YOLOv11 model, the second is a modified version, and the third is the YOLO10 and the last one is YOLOv8. Each dataset was tested with each model, resulting in a total of 24 experiments.

{\small

\begin{table*}[h!]
\centering
\caption{Performance comparison of YOLOv11, proposed versions, YOLOv10, and YOLOv8 across various datasets.}
\label{tab:model-comparison}
\begin{tabular}{|l|l|c|c|c|}
\hline
\textbf{Dataset}          & \textbf{Model}       & \textbf{Recall (\%)} & \textbf{Precision(\%)} & \textbf{mAP@50(\%)} \\ \hline
\multirow{4}{*}{WeedCrop\cite{sudars2020dataset}} 
    & YOLOv11        & 66.94 & 74.58 & 72.55 \\ 
    & YOLOv11-small  & 70.56 & 72.09 & 73.58 \\ 
    & YOLOv10        & 70.69 & 70.04 & 72.22 \\
    & YOLOv8         & 67.73 & 63.42 & 69.41 \\ \hline
\multirow{3}{*}{BCCD\cite{bccd_dataset, kutlu2020white}} 
    & YOLOv11        & 91.95 & 84.90 & 92.98 \\ 
    & YOLOv11-medium & 90.56 & 85.34 & 92.67 \\ 
    & YOLOv10        & 88.90 & 82.50 & 90.21 \\
    & YOLOv8         & 91.62 & 86.93 & 93.19 \\ \hline
\multirow{3}{*}{Underwater Pipes\cite{ciaglia2022roboflow, underwater-pipes-4ng4t_dataset}} 
    & YOLOv11        & 99.15 & 99.01 & 99.43 \\ 
    & YOLOv11-large  & 98.90 & 99.21 & 99.48 \\ 
    & YOLOv10        & 98.82 & 98.86 & 94.45 \\
    & YOLOv8         & 99.46 & 98.60 & 99.47 \\ \hline
\multirow{3}{*}{Aerial Airport\cite{ciaglia2022roboflow, aerial-airport_dataset}} 
    & YOLOv11        & 87.42 & 91.73 & 92.93 \\ 
    & YOLOv11-sm     & 86.91 & 93.15 & 93.20 \\ 
    & YOLOv10        & 86.92 & 90.18 & 92.41 \\
    & YOLOv8         & 87.05 & 90.87 & 92.23 \\ \hline
\multirow{3}{*}{Brain Tumor\cite{braintumorataset,lin2024yolov8}} 
    & YOLOv11        & 74.73 & 89.21 & 81.67 \\ 
    & YOLOv11-ml     & 72.05 & 90.90 & 80.15 \\ 
    & YOLOv10        & 73.52 & 89.23 & 80.30 \\
    & YOLOv8         & 72.46 & 90.80 & 80.49 \\ \hline
\multirow{3}{*}{Face Detection\cite{facedataset}} 
    & YOLOv11        & 93.10 & 97.98 & 96.14 \\ 
    & YOLOv11-sl     & 93.10 & 93.05 & 97.70 \\
    & YOLOv10        & 88.88 & 95.27 & 95.04 \\
    & YOLOv8         & 87.93 & 93.63 & 93.05 \\ \hline
\end{tabular}
\end{table*}}

As shown in Table \ref{tab:model-comparison}, in most scenarios, the YOLOv11 and the modified version of YOLOv11 outperforms YOLOv10 and YOLOV8 across all computed metrics, including Recall, Precision, and mAP@50. However, the performance improvement generally does not exceed 5\%. An exception is observed in the BCCD dataset, where YOLOv8 achieves slightly better performance, but the margin is less than 1\%. Regarding the performance of the original model (YOLOv11) and the proposed modified models, the results indicate that there is no significant difference in their performance. In four cases, the modified models outperform YOLOv11, as shown in the table. Specifically, YOLOv11-small, YOLOv11-large, YOLOv11-sm, and YOLOv11-sl demonstrate better performance in terms of mAP@50. Conversely, the original YOLOv11 outperforms the modified versions (YOLOv11-medium and YOLOv11-sm). Despite these variations, the maximum performance difference across all metrics is less than 2\%.

\subsection{Model Efficiency and Resource Utilization}
This section presents the performance of the modified models compared to YOLOv11, YOLOv10, and YOLOv8 in terms of  model size, GFLOPs, inference time, power consumption.

\begin{figure*}
    \centering
    \includegraphics[width=0.89\linewidth]{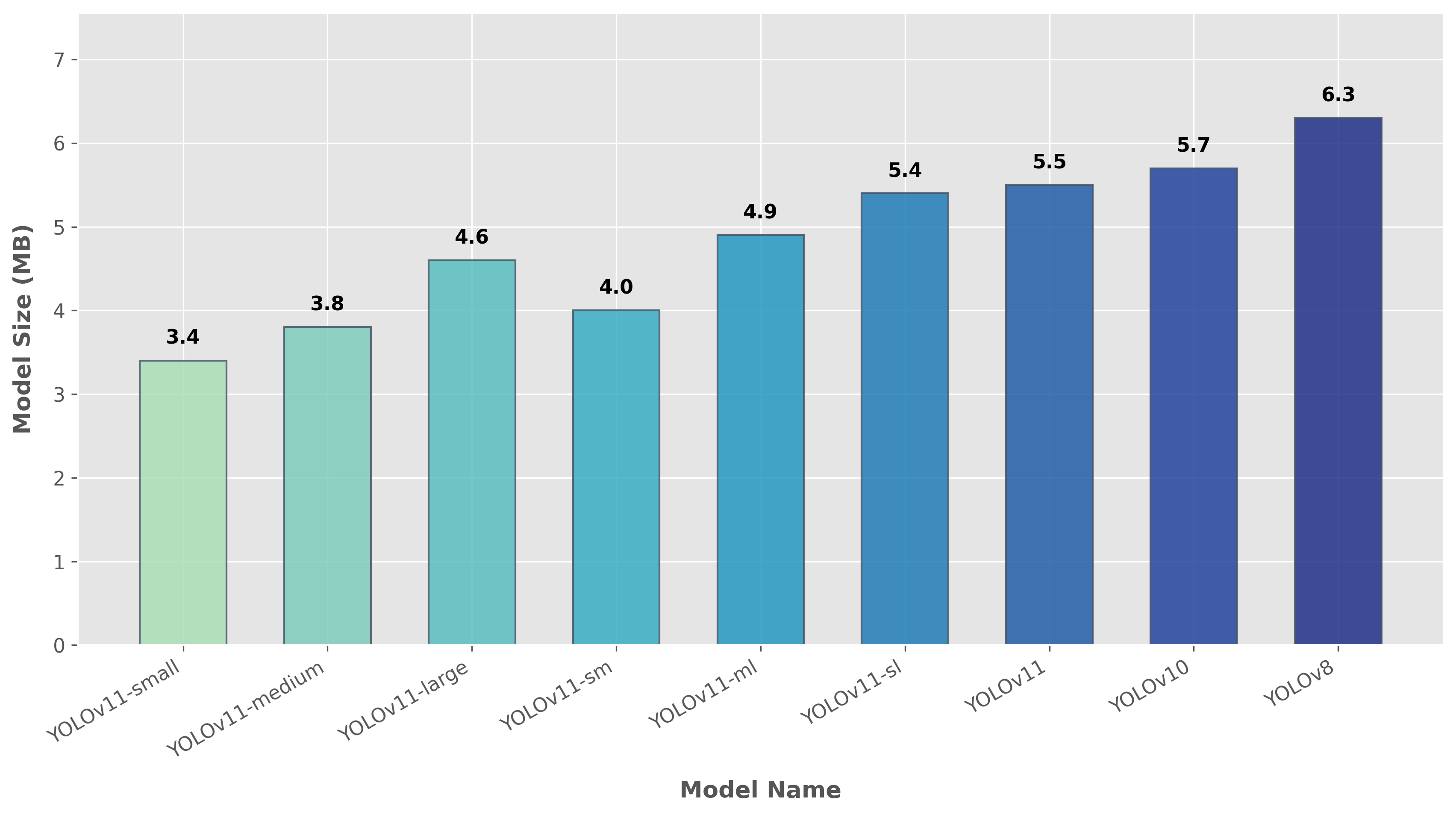}
    \caption{Model Size (MB)}
    \label{fig:model-size}
\end{figure*}

Figure \ref{fig:model-size} illustrates the sizes of the models (weights and biases) required for deployment on devices. The modified versions of YOLOv11 demonstrate significant improvements in reducing model size. As shown, all the proposed modified versions are smaller than the original YOLOv11, YOLOv10, and YOLOV8 models.

\begin{figure*}
    \centering
    \includegraphics[width=0.89\linewidth]{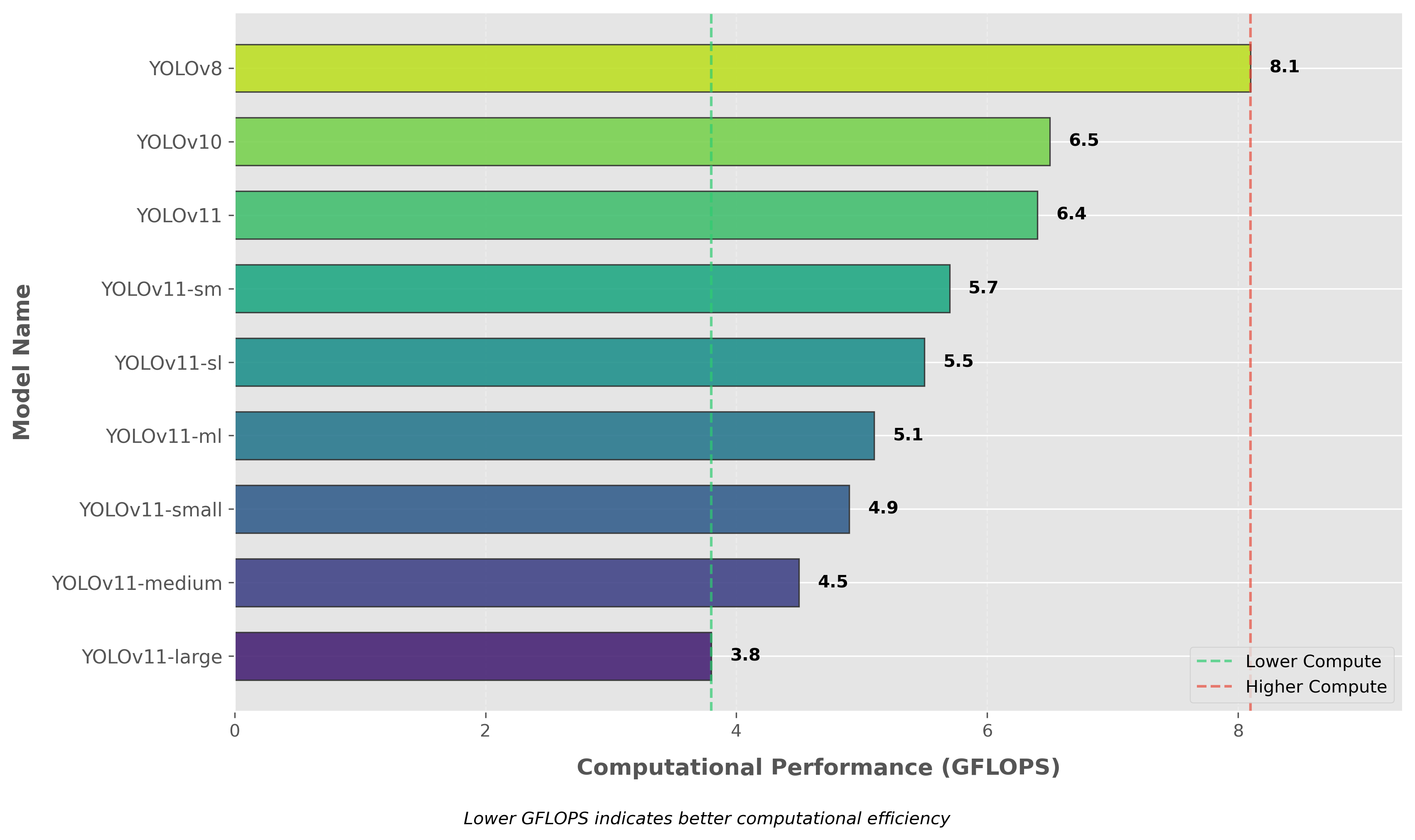}
    \caption{Computational Performance (GFLOPS)}
    \label{fig:gflops}
\end{figure*}

Regarding GFLOPS, which measures the abbreviation measures of giga floating-point operations per second, the results indicate that YOLOv10 and YOLOv8 requires higher GFLOPS compared to other models see Figure \ref{fig:gflops}. In contrast, YOLOv11 and the proposed modified versions demonstrate better efficiency. When comparing YOLOv11 to its modified versions, the results highlight a substantial reduction in GFLOPS for the modified versions.

\begin{figure*}
    \centering
    \includegraphics[width=0.89\linewidth]{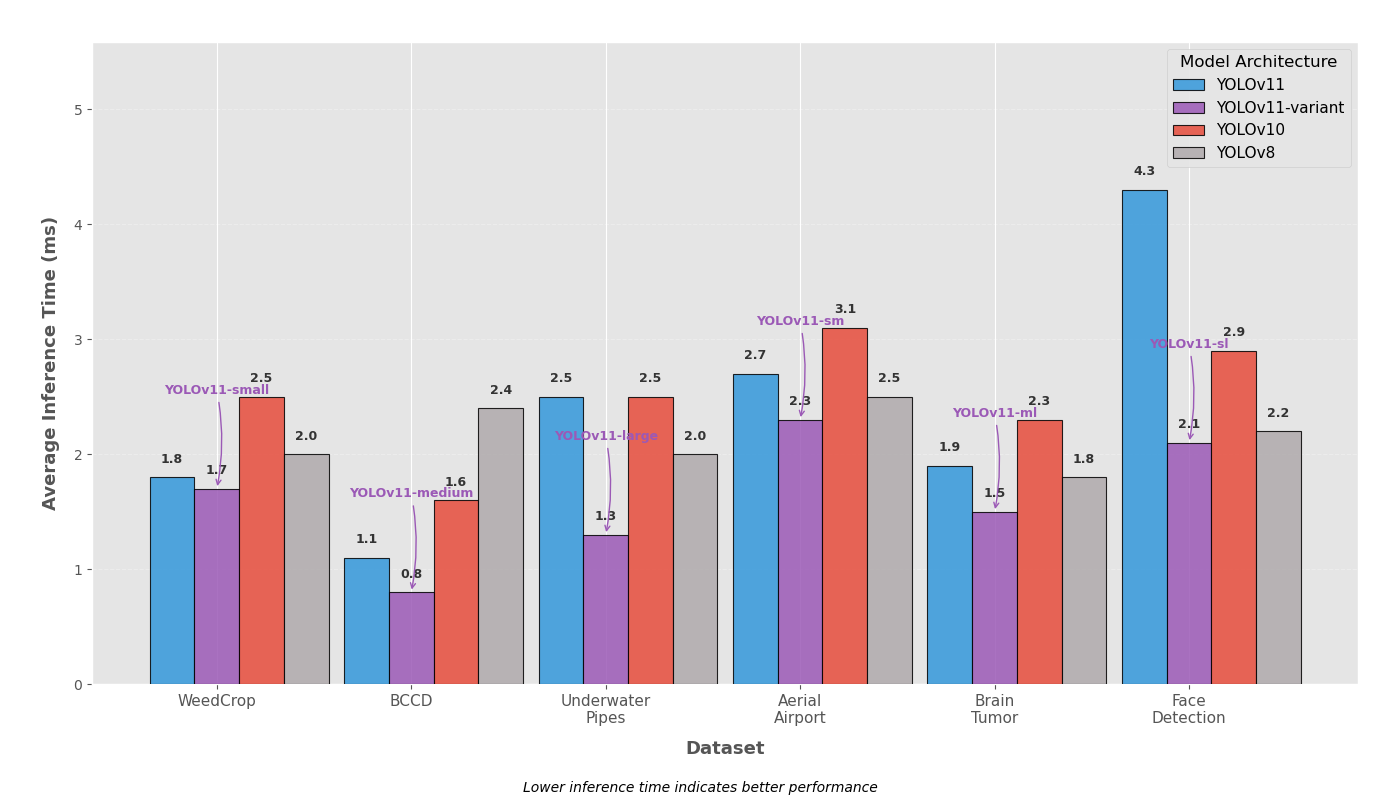}
    \caption{Average Inference Time per Image (ms)}
    \label{fig:inference-time}
\end{figure*}

The average inference time highlights the better performance of the proposed modified versions of YOLOv11 compared to the original YOLOv11, YOLOv10 and YOLOv8. As shown in Figure \ref{fig:inference-time}, the inference time of the modified versions is consistently lower across all scenarios. Additionally, all versions, including the original models, achieved an inference time of less than 5 milliseconds. The difference in inference time across all models for the same dataset is approximately 3 milliseconds. 

\begin{figure*}
    \centering
    \includegraphics[width=0.89\linewidth]{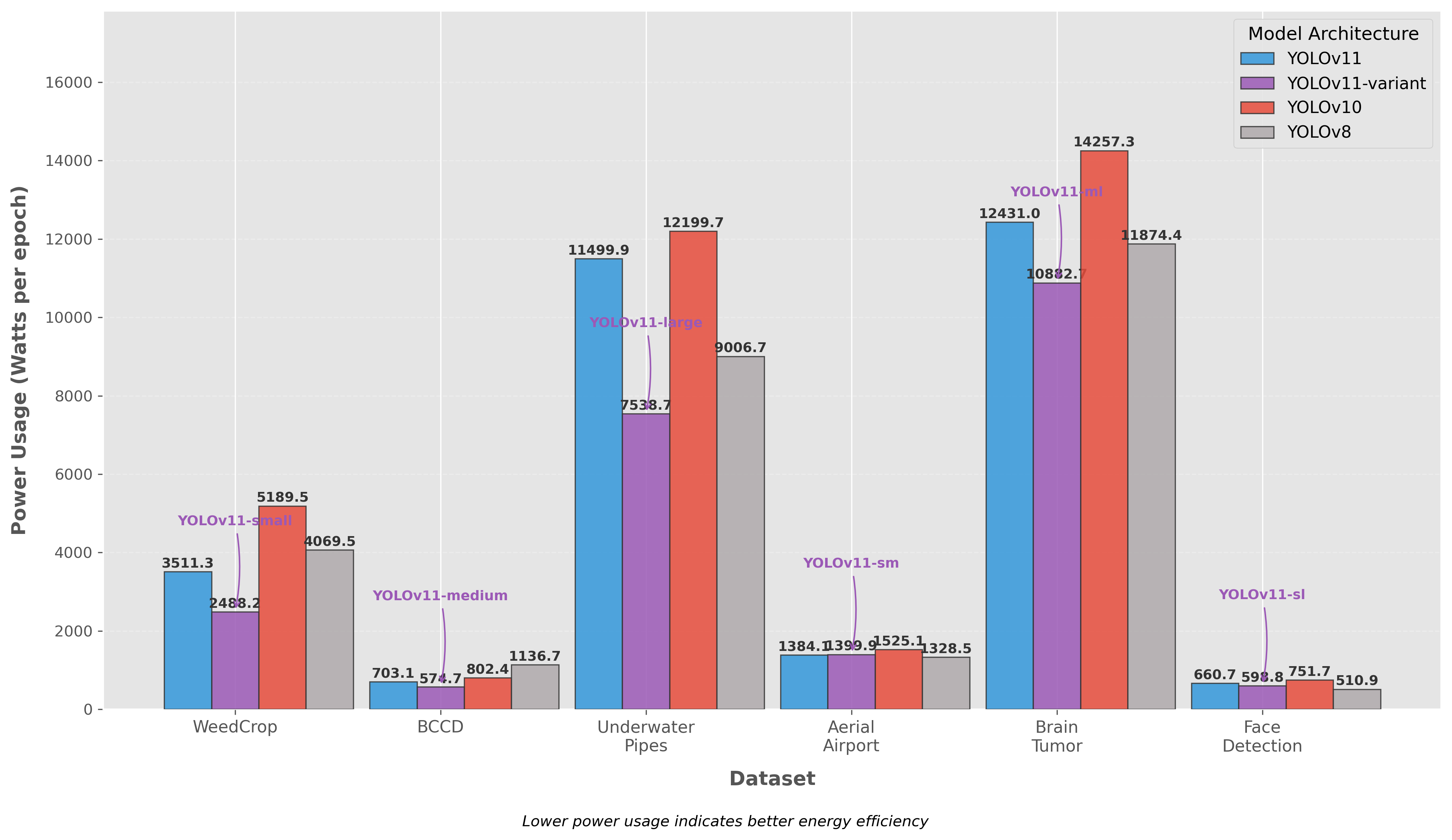}
    \caption{Average Power Usage (Watts per epoch)}
    \label{fig:power-usage}
\end{figure*}

Figure \ref{fig:power-usage} illustrates the power consumption results for each model, calculated per epoch. As shown, four of the proposed models (YOLOv11-small, YOLOv11-medium, YOLOv11-large, and YOLOv11-ml) outperform the original YOLOv11, YOLOv10, and YOLOv8 in terms of power efficiency. Meanwhile, YOLOv11-sm and YOLOv11-sl demonstrate performance comparable to the other models.

\section{Discussion}

The following the key points discussed regarding the results and how the proposed models improve performance:

\begin{itemize}
    \item [1-] Maintaining Accuracy: The proposed models were designed by removing specific blocks that do not contribute to detecting particular object sizes based on the dataset. This strategic removal of unrelated blocks has a minor impact on accuracy and, in some cases, even enhances it. For example, YOLOv11-small and YOLOv11-large perform better than the original model.

    \item [2-] Reduction in Model Size: By eliminating blocks unrelated to specific object size, the proposed models achieved size reduction. Furthermore, removing blocks associated with other object sizes contributed to additional reductions. For instance, the size of YOLOv11-sm decreased to 4 MB, and further removal of medium object-related blocks reduced it to 3.4 MB. 

    \item [3-] Reduction in GFLOPS: Removing blocks feeding the detection head that are unrelated to specific object sizes not only improved performance but also reduced the number of operations required. For example, removing both medium and large object-related components resulted in a high reduction in GFLOPS, as observed in models like YOLOv11-small. 

    \item [4-] Inference Time: Minimizing the YOLOv11 architecture resulted in a reduced number of operations required for processing, thereby decreasing the inference time. As shown in the results, all the proposed models achieved lower inference times compared to the original YOLOv11 and YOLOv8.

    \item [5-] Power Usage: The results indicate that reducing the model architecture decreases power consumption during training, this is due to the reduction in the number of operations required when the full architecture is used.  As shown, YOLOv11-small, YOLOv11-medium, YOLOv11-large, and YOLOv11-ml achieved lower power consumption compared to both the original YOLOv11, YOLOv10, and YOLOv8. Meanwhile, YOLOv11-sm and YOLOv11-sl demonstrated approximately the same power consumption as the original YOLOv11 and YOLOv8.
\end{itemize}

\section{Limition and Future works}

Although the proposed modified versions of YOLOv11 performed well in most performance metrics, such as power consumption, reduction of model size, inference time, and computational efficiency, their suitability depends on the target application. For instance, in medical applications where object sizes (e.g., cells) remain constant, the modified versions may be ideal. However, for applications involving objects with varying sizes, such as car detection on roads where cars appear larger when closer to the camera and smaller when farther away, using the full version of YOLOv11 might be more effective.

To address such scenarios, we recommend implementing a preprocessing step to classify dataset objects by size using the program which is mentioned in Section 4.1, enabling the selection of the most suitable modified version. In addition, we suggest testing the proposed models in uncontrolled environments, such as foggy conditions.

Furthermore, the current implementation uses 32-bit integers to represent model parameters. We recommend experimenting with different quantization methods to reduce the model size and evaluating the models both before and after applying quantization techniques.

\section{Conclusion}
This paper presents six modified versions based on the latest YOLOv model, YOLOv11. These modified versions are designed to target the detection of specific object size ranges. Each of the six proposed models focuses on a particular object size category, such as small, medium, large, small-medium, medium-large, and small-large objects.

To select the appropriate model for a given dataset, an analysis of the dataset is required to determine the predominant object sizes. For this purpose, we developed a program to compute and classify object sizes effectively. Each proposed model was tested on different datasets and compared with two other models, YOLOv11 and YOLOv8.

The overall results demonstrate that using the modified versions instead of training on whole model (YOLOv11, or YOLOv8) to detect all object sizes provides better performance. These modified versions also utilize fewer resources compared to the original YOLOv11 and YOLOv8 models. Despite the reduced resource usage, the accuracy of the modified models did not decrease significantly compared to the original models. In some cases, the modified models even outperformed the original versions. Ultimately, we also mention the main limitations of the work and suggest different areas that can be addressed in the future.

{\small
\textbf{Author‘s contribution}
Areeg Fahad Rasheed  carried out the implementation and writing along with M. Zarkoosh.

\small
\subsubsection*{Data availability}
The datasets mentioned in this paper are available through the cited sources.

\small
\section*{Declarations}
\small
\subsection*{Conflict of interest}
 The authors declare that they have no conflict of
interest

\small
\subsection*{Ethical approval}
Not applicable

\small
\section*{Code Availability}
The code used for the analysis and experiments in this study will be made available upon request and publication. 
}

\bibliography{refs}
\bibliographystyle{abbrv}

\end{document}